\documentclass{article}

\PassOptionsToPackage{numbers, compress}{natbib}
\usepackage[preprint]{neurips_2026}

\usepackage[utf8]{inputenc} 
\usepackage[T1]{fontenc}    
\usepackage{hyperref}       
\usepackage{url}            
\usepackage{booktabs}       
\usepackage{amsfonts}       
\usepackage{nicefrac}       
\usepackage{microtype}      
\usepackage{xcolor}         

\usepackage{soul}

\usepackage{threeparttable}
\usepackage{float}
\usepackage{tabularx}

\usepackage{graphicx}

\usepackage{tikz}
\usetikzlibrary{positioning}

\usepackage{titlesec}
\titlespacing*{\subsection}
  {0pt}    
  {6pt}    
  {4pt}    

\usepackage{titlesec}
\titlespacing*{\paragraph}
{0pt}   
{0.5em} 
{0.5em} 

\usepackage{caption}

\usepackage{enumitem}
\setlist[itemize]{leftmargin=*, itemsep=1pt, topsep=2pt, parsep=0pt, partopsep=0pt}

\usepackage{booktabs}
\usepackage{threeparttable}
\usepackage{pifont}
\usepackage{amssymb}
\newcommand{\cmark}{\ding{51}}        
\newcommand{\xmark}{\ding{55}}        
\newcommand{\pmark}{$\checkmark^{*}$} 

\usepackage[table]{xcolor}
\definecolor{oursblue}{RGB}{232,242,250}    
\usepackage{adjustbox}

\usepackage{amsmath}

\usepackage{subcaption}

\usepackage{cleveref}

\usepackage{booktabs}
\usepackage{tabularx}
\usepackage[table]{xcolor}
\usepackage{array}

\hypersetup{
    colorlinks=true,
    linkcolor=red,
    citecolor=blue,
    urlcolor=blue
}

\crefname{figure}{Fig.}{Figs.}
\Crefname{figure}{Figure}{Figures}

\crefname{table}{Table}{Tables}
\Crefname{table}{Table}{Tables}

\crefname{equation}{Eq.}{Eqs.}
\Crefname{equation}{Equation}{Equations}
\creflabelformat{equation}{#2(#1)#3}

\crefname{section}{Section}{Sections}
\Crefname{section}{Section}{Sections}

\crefname{subsection}{Section}{Sections}
\Crefname{subsection}{Section}{Sections}

\title{TextRich: A Multi-Domain Benchmark for Detecting AI-Generated Text-Rich Images from GPT-Image-2}

\author{%
Yijin Wang\textsuperscript{$*$} \quad
Shuyi Wang\textsuperscript{$*$} \quad
Wenhan Zhang \quad
Yuqi Ouyang\textsuperscript{$\dagger$}\\[1mm]
College of Computer Science, Sichuan University\\
}

\begin{document}

\maketitle

\begingroup
\renewcommand{\thefootnote}{}
\footnotetext{$*$: Equal contribution}
\footnotetext{$\dagger$: Correspondence at yuqi.ouyang@scu.edu.cn}
\endgroup

\begin{abstract}
  Text-rich images often contain privacy-sensitive, transactional, or decision-relevant information. As recent multimodal image generation models become increasingly capable of synthesizing realistic textual content and structured visual designs, detecting AI-generated text-rich images has become an important challenge for digital trust and content authenticity. Existing benchmarks, however, largely focus on object-centric images and provide limited coverage of scenarios where textual semantics and layout organization are central. In this paper, we introduce TextRich, a multi-domain benchmark for detecting text-rich images generated by OpenAI's GPT-Image-2. The benchmark contains 12,095 images across six representative categories: commercial posters, infographic charts, academic posters, receipts, tables, and UI screenshots. Using this benchmark, we evaluate five representative AI-generated image detectors under a zero-shot setting and further explore the capability of a multimodal vision-language model for this task. Our results reveal substantial performance variations across text-rich domains, where existing AI-generated image detectors exhibit distinct strengths and failure modes. Although the strongest detector achieves competitive overall performance, it remains ineffective on certain structured categories and highly sensitive to JPEG compression. Vision-language models provide a promising complementary approach, but still struggle with highly structured text-rich images. These findings highlight the need for text- and layout-aware detection methods for modern AI-generated images. Our dataset is released at \url{https://huggingface.co/datasets/Shuyiww/TextRich}.

\end{abstract}

\section{Introduction}

Generating realistic textual content has long been a challenging problem for image generation models. Earlier systems frequently produced distorted characters, inconsistent word structures, and semantically meaningless text, making AI-generated images relatively easy to identify in text-rich scenarios. Recent advances in multimodal image generation have substantially improved this capability~\citep{ICLR2024_081b0806, NEURIPS2023_1df4afb0}. Modern image generation models are now able to render coherent and meaningful text while maintaining consistency with surrounding visual content.

The ability to generate high-quality text has expanded the scope of AI-generated images beyond traditional artistic and creative applications. Today, image generation systems can produce a wide range of text-rich content, including User Interfaces (UI screenshots), receipts, tables, posters, and other document-like images. As these images become increasingly realistic and accessible, concerns have emerged regarding the creation of misleading or authentic-looking content~\citep{wu2026aiforgedocbenchmarkdetectingaiforged}. Consequently, reliable image authenticity assessment is becoming increasingly important in text-rich scenarios.

A recent example of this trend is OpenAI's GPT Image 2, hereafter referred to as GPT-Image-2, which demonstrates strong capability in generating images with readable text and structured layouts~\citep{OpenAI2026Images2}. Images generated by such systems can closely resemble real-world visual documents, creating new challenges for AI-generated image (AIGI) detection. Compared with conventional object-oriented images, text-rich images often involve textual information, spatial arrangements, and structured layouts that jointly contribute to image semantics, making authenticity assessment more challenging.

Despite the growing importance of this problem, existing datasets provide limited support for studying text-rich AI-generated images~\citep{NEURIPS2023_f4d4a021, hong2024wildfakelargescalechallengingdataset, 10222083}. Existing AIGI datasets primarily focus on visual content and contain little meaningful text, while existing text-related datasets are often restricted to specific application domains~\citep{wu2026aiforgedocbenchmarkdetectingaiforged, zhang2026gpt4oreceiptdatasethumanstudy, hu2026scifigdetectbenchmarkaigeneratedscientific}. Moreover, relatively few datasets are constructed using recent multimodal image generation models such as GPT-Image-2. As a result, there remains a lack of datasets that simultaneously cover diverse text-rich scenarios and modern image generation systems. In addition, the performance of existing AIGI detectors on such content has not been systematically evaluated.

To address these gaps, we introduce TextRich, a new multi-domain dataset of text-rich images generated using GPT-Image-2, covering six representative categories: UI Screenshot, Receipt, Table, Infographic Charts, Academic Poster, and Commercial Poster. Based on this dataset, we establish a benchmark for text-rich AIGI detection and conduct a unified evaluation of several representative detection methods across different scenarios. The main contributions of this work are as follows:
\begin{itemize}
    \item We introduce a multi-domain benchmark of GPT-Image-2 generated text-rich images covering six representative categories, enabling systematic evaluation of modern multimodal image generation.

    \item We propose a two-dimensional formulation of text-rich image generation based on text-layout regularity and functional context, together with a privacy-preserving prompt synthesis pipeline that distills structural, semantic, and stylistic cues from real data to enable controlled generation without using original text or identities.

    \item We establish a unified benchmark for AI-generated text-rich image detection, with comprehensive evaluations across diverse detection paradigms that reveal category-dependent performance variations, limited robustness under common post-processing, and the remaining challenges of vision-language models on structured text-rich images.
    


\end{itemize}

\section{Related Work}
\label{sec:related_work}
\subsection{AI-Generated Image Datasets}

AIGI datasets have become an important foundation for developing and evaluating image authenticity detectors. Representative datasets such as GenImage, ArtiFact, and WildFake contain images generated by a variety of image generation models, including Generative Adversarial Networks (GANs), Diffusion Models (DMs), and other earlier image generation frameworks~\citep{NEURIPS2023_f4d4a021, 10222083, hong2024wildfakelargescalechallengingdataset}. These datasets provide large-scale resources for studying AIGI detection and detector generalization. However, they mainly focus on object-oriented visual content and contain little or no meaningful textual information. Moreover, most generated images in these datasets originate from earlier generation models and provide limited coverage of recent multimodal image generators.


Recognizing the growing importance of textual content in AI-generated images, several datasets have been proposed for text-related forgery detection. Tampered-IC13, OSTF, and AIForge-Doc focus on localized text manipulation in scene-text and document-oriented images~\citep{10.1007/978-3-031-19815-1_13, qu2025revisiting, wu2026aiforgedocbenchmarkdetectingaiforged}. In contrast, GPT4o-Receipt and SciFigDetect consider fully AI-generated receipts and scientific figures, reflecting a shift toward synthetic text-rich image generation by modern image generation systems~\citep{zhang2026gpt4oreceiptdatasethumanstudy, hu2026scifigdetectbenchmarkaigeneratedscientific}.

Despite these advances, existing datasets remain fragmented across individual application domains and provide limited coverage of diverse text-rich scenarios~\citep{ding2026deep, fu2025multimodal}. As a result, important categories such as academic posters, commercial posters, tables, and mobile user interfaces remain underrepresented. Moreover, most existing datasets are built using earlier generative models and do not reflect the capabilities of recent multimodal image generation systems such as GPT-Image-2, which can produce highly coherent text and structured layouts. Our dataset complements existing resources by covering multiple representative text-rich image categories generated by a GPT-Image-2.

\begin{figure}[!t]
    \centering
    \includegraphics[width=\linewidth]{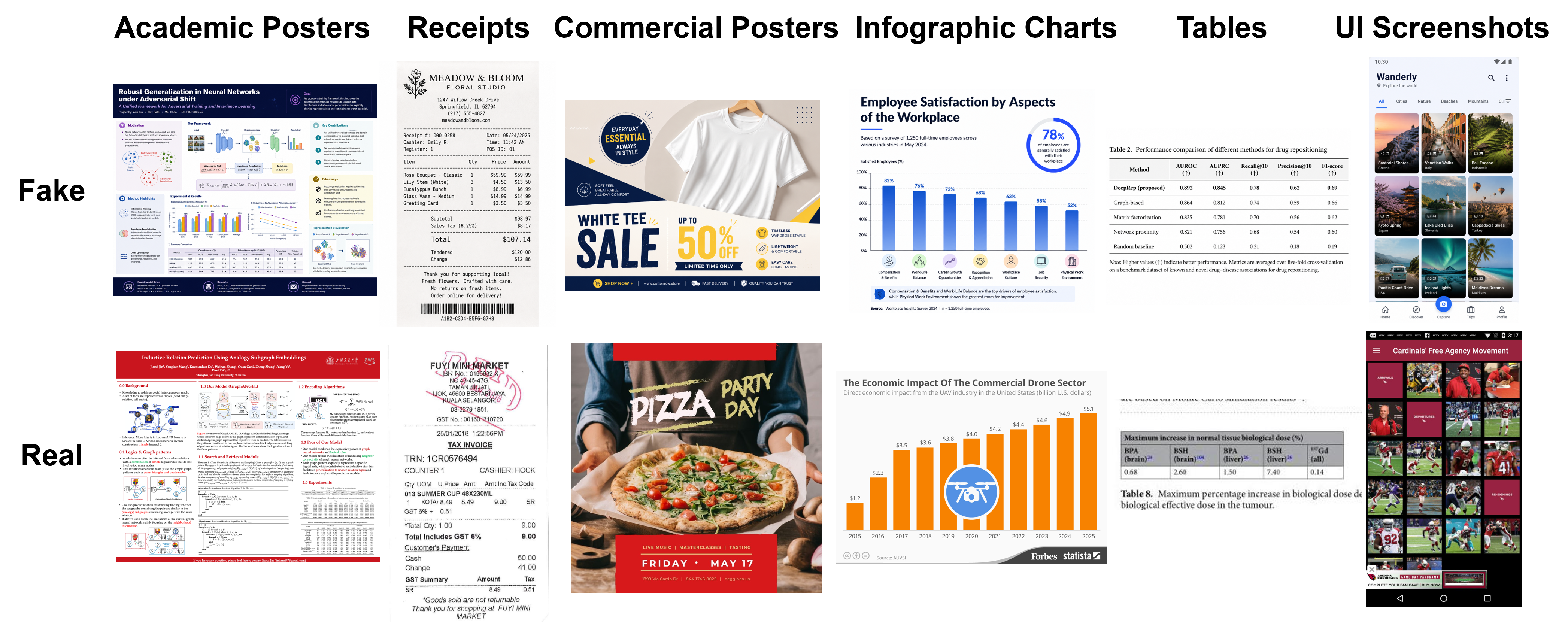}
    \caption{Visual examples from TextRich, real and synthetic images for all six text-rich categories.}
    \label{fig:dataset_sample_overview}
\end{figure}

\renewcommand{\arraystretch}{1.1}
\begin{table}[!t]
\centering
\begin{threeparttable}
\small
\setlength{\tabcolsep}{3pt}

\begin{tabularx}{\textwidth}{@{}l*{7}{>{\centering\arraybackslash}X}@{}}
\toprule
\textbf{Category} 
& \textbf{Tables} 
& \textbf{Receipts} 
& \textbf{Infographic Charts} 
& \textbf{UI Screenshots} 
& \textbf{Comm. Posters} 
& \textbf{Acad. Posters} 
& \textbf{Total} \\
\midrule
Generated & 1000 & 1095 & 1000 & 1000 & 1010 & 1004 & 6109 \\
Real      & 1000 & 986 & 1000 & 1000 & 1000  & 1000  & 5986 \\
\midrule
Total     & 2000 & 2081 & 2000 & 2000 & 2010 & 2004 & 12095 \\
\bottomrule
\end{tabularx}

\caption{Category-wise number of real and generated images in TextRich. Comm. Posters and Acad. Posters denote commercial posters and academic posters, respectively.}
\label{tab:dataset_statistics_overview}
\end{threeparttable}
\end{table}

\subsection{AI-Generated Image Detection}
AIGI detection has been widely studied by exploiting different visual cues, including generation artifacts, statistical patterns, and learned representations. Early methods primarily focused on spatial artifacts introduced by CNN-based generative models, with CNNSpot~\citep{wang2020cnngenerated} demonstrating that detectors trained on one generator can generalize to unseen architectures. With the emergence of diffusion models, subsequent approaches explored more generalizable cues beyond model-specific artifacts. NPR~\citep{tan2024rethinking} captures local neighboring pixel relationships caused by generative up-sampling operations, while frequency-based methods such as FreqNet~\citep{tan2024frequency} and SPAI~\citep{karageorgiou2025any} exploit spectral characteristics to improve cross-generator robustness. More recently, vision foundation models have been introduced for AIGI detection. UnivFD~\citep{ojha2024universalfakeimagedetectors} leverages pretrained visual representations with lightweight classifiers, achieving strong generalization without requiring forgery-specific training.

Although these methods have achieved promising results on existing AIGI benchmarks, their effectiveness on text-rich images generated by recent multimodal image generation systems remains largely unexplored. Since text-rich images involve complex interactions among textual semantics, spatial layouts, and visual structures, evaluating existing detectors in this setting provides valuable insights into their generalization ability beyond conventional natural images.

\section{Dataset Construction}
\subsection{Dataset Overview}


To characterize real-world text-rich images, we organize text-rich images along two dimensions. The first dimension is the degree of text-layout regularity, ranging from free-form visual composition to highly structured textual arrangements. The second dimension is the functional context in which the image is used, including visual communication, information presentation, academic communication, transaction records, structured data representation, and human-computer interaction. These contexts are also closely related to practical risks, where generated or manipulated text-rich images may affect information trustworthiness, privacy-sensitive records, and user-facing interfaces.

Based on these considerations, we include six representative categories in our dataset: commercial posters, infographic charts, academic posters, receipts, tables, and UI screenshots. 
Representative examples of the six categories are shown in \cref{fig:dataset_sample_overview}.

In addition, \cref{tab:dataset_statistics_overview} summarizes the number of real and generated images in each category. Overall, the dataset contains 12,095 images, with 6,109 generated and 5,986 real samples. Each category contains approximately 2,000 images, with roughly half randomly sampled from the corresponding source dataset and the other half generated using our pipeline.




\begin{figure}[!t]
    \centering
    \includegraphics[width=\linewidth]{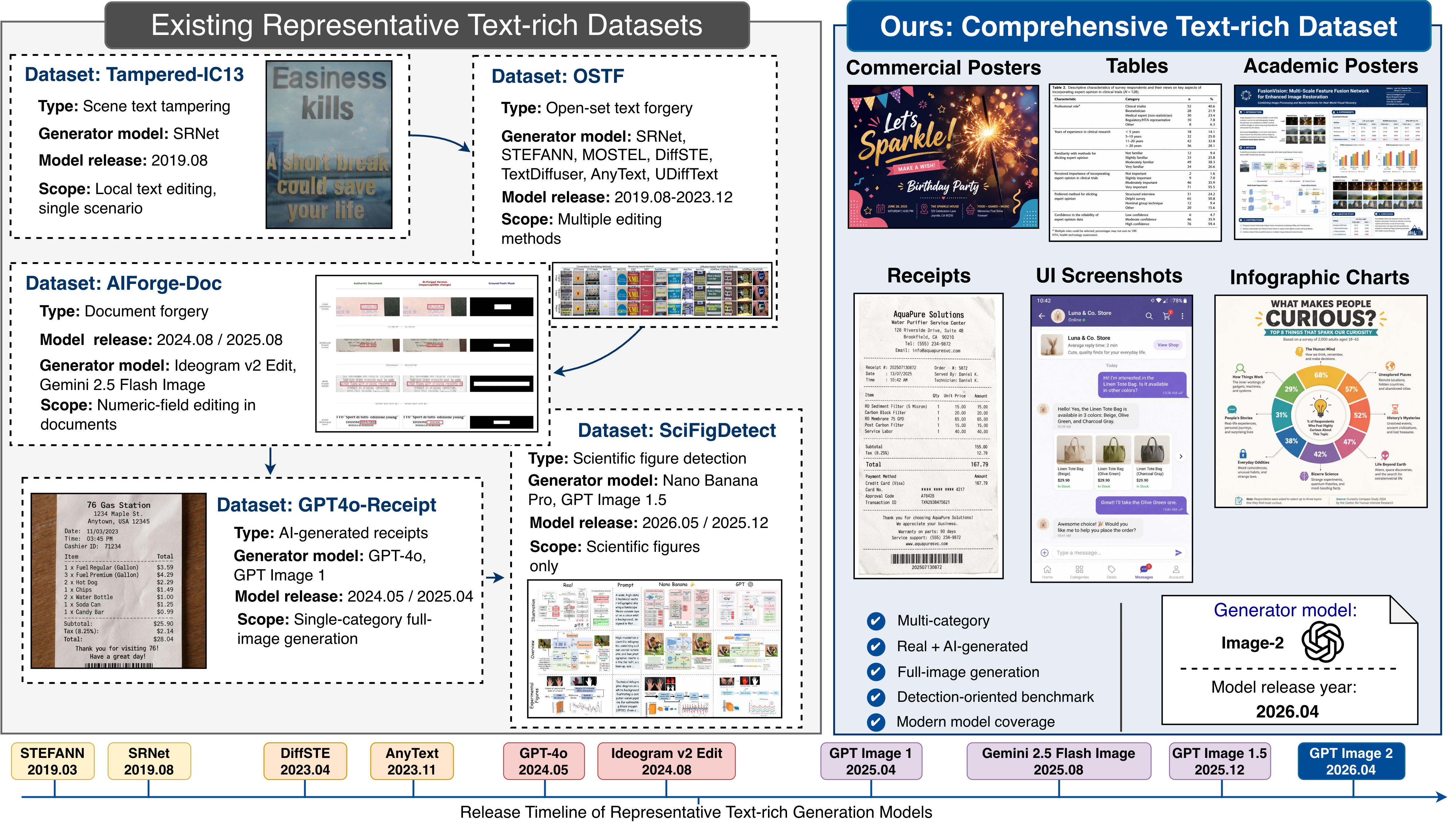}
    \caption{Comparison of representative text-rich datasets and TextRich.}
    \label{fig:dataset_comp}
\end{figure}

\begin{table}[!t]
\centering
\small
\setlength{\tabcolsep}{4pt}
\begin{tabular*}{\textwidth}{@{\extracolsep{\fill}}lccccc@{}}
\toprule
\textbf{Dataset} 
& \textbf{Multi-cat.} 
& \textbf{Real Class} 
& \textbf{Gen./Forged} 
& \textbf{Full-image Gen.} 
& \textbf{Detection} \\
\midrule
Tampered-IC13~\citep{10.1007/978-3-031-19815-1_13}
& \xmark 
& \xmark 
& \cmark 
& \xmark 
& \cmark \\

OSTF~\citep{qu2025revisiting}
& \pmark 
& \xmark 
& \cmark 
& \xmark 
& \cmark \\

AIForge-Doc~\citep{wu2026aiforgedocbenchmarkdetectingaiforged}
& \pmark 
& \cmark 
& \cmark 
& \xmark 
& \cmark \\

GPT4o-Receipt~\citep{zhang2026gpt4oreceiptdatasethumanstudy}
& \xmark 
& \cmark 
& \cmark 
& \cmark 
& \cmark \\

SciFigDetect~\citep{hu2026scifigdetectbenchmarkaigeneratedscientific}
& \pmark 
& \cmark 
& \cmark 
& \cmark 
& \cmark \\

\rowcolor{oursblue}[0pt][0pt]
\textbf{Ours} 
& \cmark 
& \cmark 
& \cmark 
& \cmark 
& \cmark \\
\bottomrule
\end{tabular*}
\caption{Comparison between existing text-rich datasets and TextRich. \pmark{} indicates partial support.}
\label{tab:existing_dataset_comparison}
\end{table}

\subsection{Comparison with Existing Datasets}
As AI systems become increasingly capable of generating images with embedded textual content, text-rich datasets are becoming important for AIGI detection and for evaluating future multimodal models in text-image generation scenarios. In this setting, a useful benchmark should reflect the complexity of real-world text-rich visual content. Specifically, it should cover multiple representative categories with diverse layouts and practical visual formats, include both real and generated samples for detector evaluation, and consider full-image generation, where the entire image is synthesized rather than only local text regions being modified.

Based on these considerations, we compare our benchmark with existing text-rich datasets in \cref{tab:existing_dataset_comparison} and \cref{fig:dataset_comp}. Existing datasets have made valuable contributions to text-rich image understanding and AI-generated text-rich image forensics. However, many of them were not originally designed to jointly cover multiple text-rich categories, real and generated samples, and full-image generation settings. Our benchmark complements these resources by providing a unified multi-category evaluation setting for detecting AI-generated text-rich images.

\begin{figure}[!t]
    \centering
    \includegraphics[width=\linewidth]{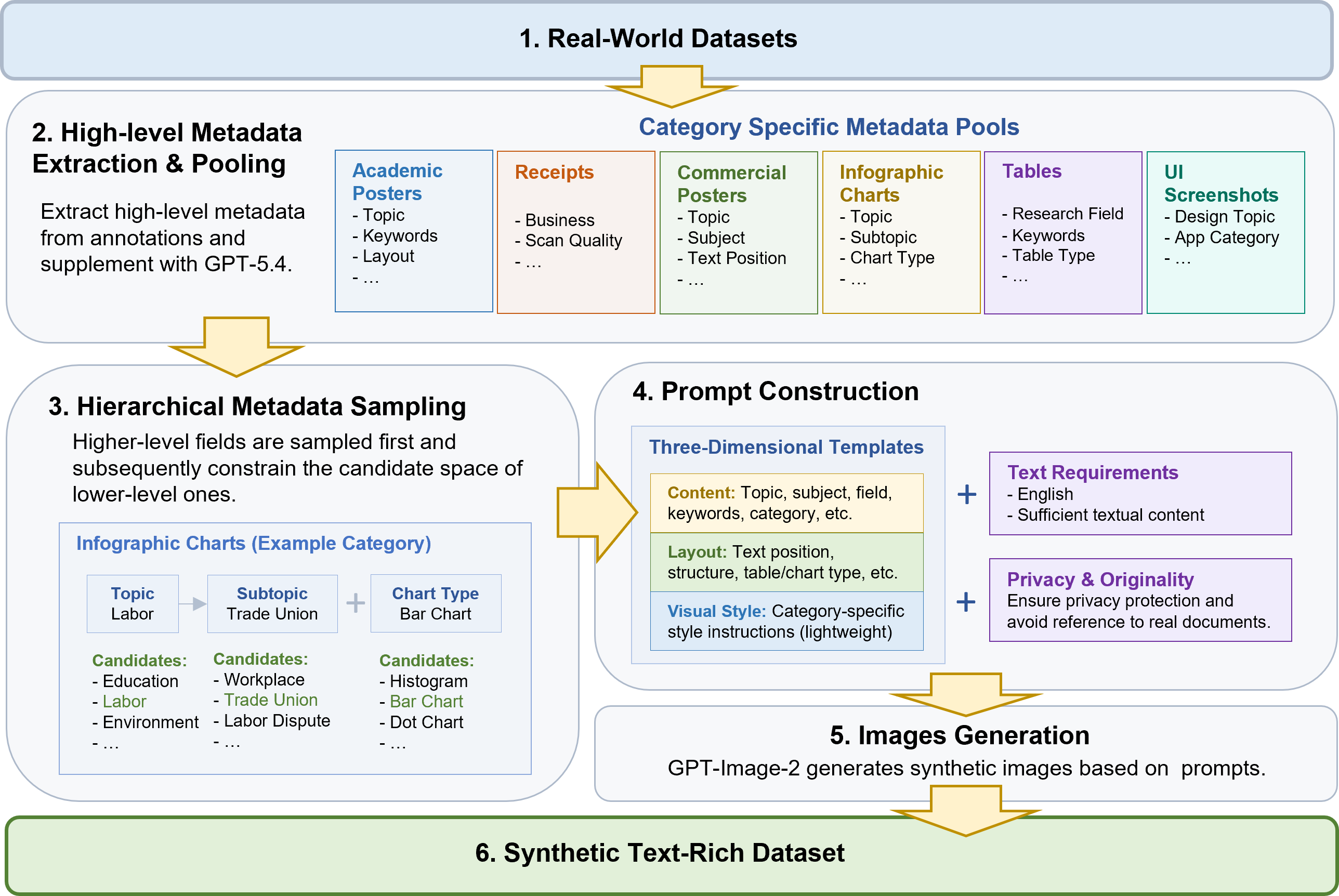}
    \caption{Overview of the construction pipeline of TextRich.}
    \label{fig:construction_pipeline}
\end{figure}

\subsection{Dataset Construction}

To construct the benchmark, we first collect real images from existing public datasets for each target category, and then generate the corresponding synthetic images through a data-guided prompt generation strategy. Specifically, the real split covers six text-rich image domains: commercial posters, infographic charts, academic posters, receipts, tables, and UI screenshots. For each domain, we select a representative real-world dataset as the source of real samples, including Crello~\citep{Yamaguchi_2021_ICCV} for commercial posters, ChartGalaxy~\citep{li2025chartgalaxy} for infographic charts, PosterSum~\citep{saxena2025postersum} for academic posters, SROIE~\citep{huang2019icdar2019} for receipts, PubTables-1M~\citep{smock2022pubtables} for tables, and Enrico~\citep{leiva2020enrico} for UI screenshots. These datasets are chosen because they cover diverse forms of text-rich visual content, ranging from structured documents to flexible graphic designs. The collected real samples serve not only as the real portion of the benchmark, but also as domain references for constructing prompts for synthetic data construction. 

Depicted in \cref{fig:construction_pipeline}, synthetic images are generated using prompts derived from real-world datasets. For each category, we first identify high-level metadata fields that capture the key characteristics of the source data while avoiding direct replication of individual samples. These metadata are extracted from dataset annotations and further complemented by image analysis using GPT-5.4~\citep{openai2026gpt54}, and then organized into category-specific pools. Next, during prompt construction, metadata configurations are sampled hierarchically from the corresponding pool, where higher-level attributes constrain the selection of lower-level ones to ensure semantic consistency. The sampled configurations are incorporated into category-specific prompt templates structured around three dimensions: content, layout, and visual style. While metadata primarily control the content and layout of generated images, lightweight style instructions are introduced to maintain category-level visual consistency while preserving diversity. All templates additionally include shared requirements for text as well as privacy and originality, which enforce English text generation, sufficient textual content, avoidance of real personal information, and prevention of direct reproduction of existing images, documents, or publications. As a result, this pipeline enables the generation of diverse text-rich images that retain the structural characteristics of real-world datasets without relying on individual samples, while reducing privacy risks and potential content duplication.


\section{Experiments}
Having constructed the proposed dataset with six text-rich categories, we proceed to evaluate existing AIGI detectors. We select a set of representative detection methods to benchmark performance, and analyze the results to understand their strengths, limitations, and areas for improvement in handling text-rich visual domains.

\subsection{Baseline Methods}
For our benchmark, we evaluate six representative approaches from both specialized AIGI detectors and general-purpose multimodal models. The first group consists of five AIGI detectors from open-source repositories and academic studies, covering diverse detection paradigms including artifact-based analysis, pixel-level pattern modeling, frequency-domain characterization, and foundation-model-based representations. In addition, we include Qwen3.5-27B as an exploratory vision-language baseline to investigate the capability of general multimodal reasoning for detecting AI-generated text-rich images.

\paragraph{CNNSpot.}
CNNSpot is a CNN-based AIGI detector trained to identify images produced by CNN-based generative models~\citep{wang2020cnngenerated}. Its key finding is that a classifier trained on images from one generator can generalize to images from unseen generator architectures when appropriate pre-processing and data augmentation are used. This makes CNNSpot a representative artifact-based detector for cross-generator evaluation.

\paragraph{NPR.}
NPR is based on Neighboring Pixel Relationships, a local artifact representation designed for generalizable synthetic-image detection~\citep{tan2024rethinking}. It captures pixel-level dependencies introduced by up-sampling operations in generative pipelines. By focusing on local up-sampling traces rather than high-level semantics, NPR is designed to generalize across unseen GAN and diffusion models.

\paragraph{FreqNet.}
FreqNet is a lightweight frequency-aware detector designed for cross-generator generalizability~\citep{tan2024frequency}. It extracts source-agnostic representations by processing both phase and amplitude spectra in the Fourier domain alongside high-frequency spatial features. By avoiding overfitting to model-specific artifacts, FreqNet serves as a competitive frequency-domain baseline.

\paragraph{SPAI.}
SPAI addresses cross-generator generalization by using the spectral distribution of real images as an invariant discriminative pattern~\citep{karageorgiou2025any}. It trains a self-supervised model via masked frequency reconstruction on real data, identifying synthetic images through spectral reconstruction similarity and spectral context attention. This enables SPAI to reliably capture subtle spectral inconsistencies across unseen generative architectures.

\paragraph{UnivFD.}
To mitigate classifier bias toward specific forgery patterns, UnivFD leverages feature spaces from large pretrained vision-language models~\citep{ojha2024universalfakeimagedetectors}. It replaces end-to-end classification training with simple linear probing or nearest neighbor search over these representations. This approach demonstrates strong generalization across unseen diffusion and autoregressive architectures.

\paragraph{Qwen3.5-27B.}
In addition to specialized AIGI detectors, we include Qwen3.5-27B as an exploratory multimodal baseline. Unlike conventional detectors trained specifically for AIGI detection, Qwen3.5-27B is a general-purpose vision-language model capable of jointly understanding visual and textual information~\citep{qwen3.5}. We evaluate its zero-shot capability by prompting the model to classify each image as either real or AI-generated, providing an analysis of whether general multimodal reasoning can complement specialized forensic approaches in text-rich scenarios.

Overall, these baselines cover complementary detection paradigms, ranging from low-level forensic cues to high-level multimodal understanding. This diversity enables a comprehensive evaluation of how existing AIGI detection approaches generalize to text-rich visual domains beyond conventional natural images.

\subsection{Implementation Details}
All detectors were evaluated using publicly available pretrained checkpoints obtained from their official releases. No detector was fine-tuned, retrained, or otherwise adapted to our dataset. All

\begin{minipage}[b]{0.48\textwidth}
    models were executed using their official inference pipelines under a zero-shot evaluation setting. For the multimodal evaluation, Qwen3.5-27B was accessed through its official API. Input preprocessing followed the detector-specific procedures provided in the official implementations. Since different detectors rely on different backbone architectures and preprocessing pipelines, image resizing, normalization, and input formatting were handled by the corresponding official codebases. No additional preprocessing, manual cropping, or dataset-specific image transformations were applied. Furthermore, no test-time augmentation was used. For robustness evaluation, we additionally generated degraded versions of the images following the protocol in GenImage~\citep{NEURIPS2023_f4d4a021}, including JPEG compression with a quality factor of 65, resolution reduction to $112 \times 112$, and Gaussian blur with $\sigma=3$.
\end{minipage}
\hfill
\begin{minipage}[b]{0.48\textwidth}
    \centering
    \includegraphics[width=\textwidth]{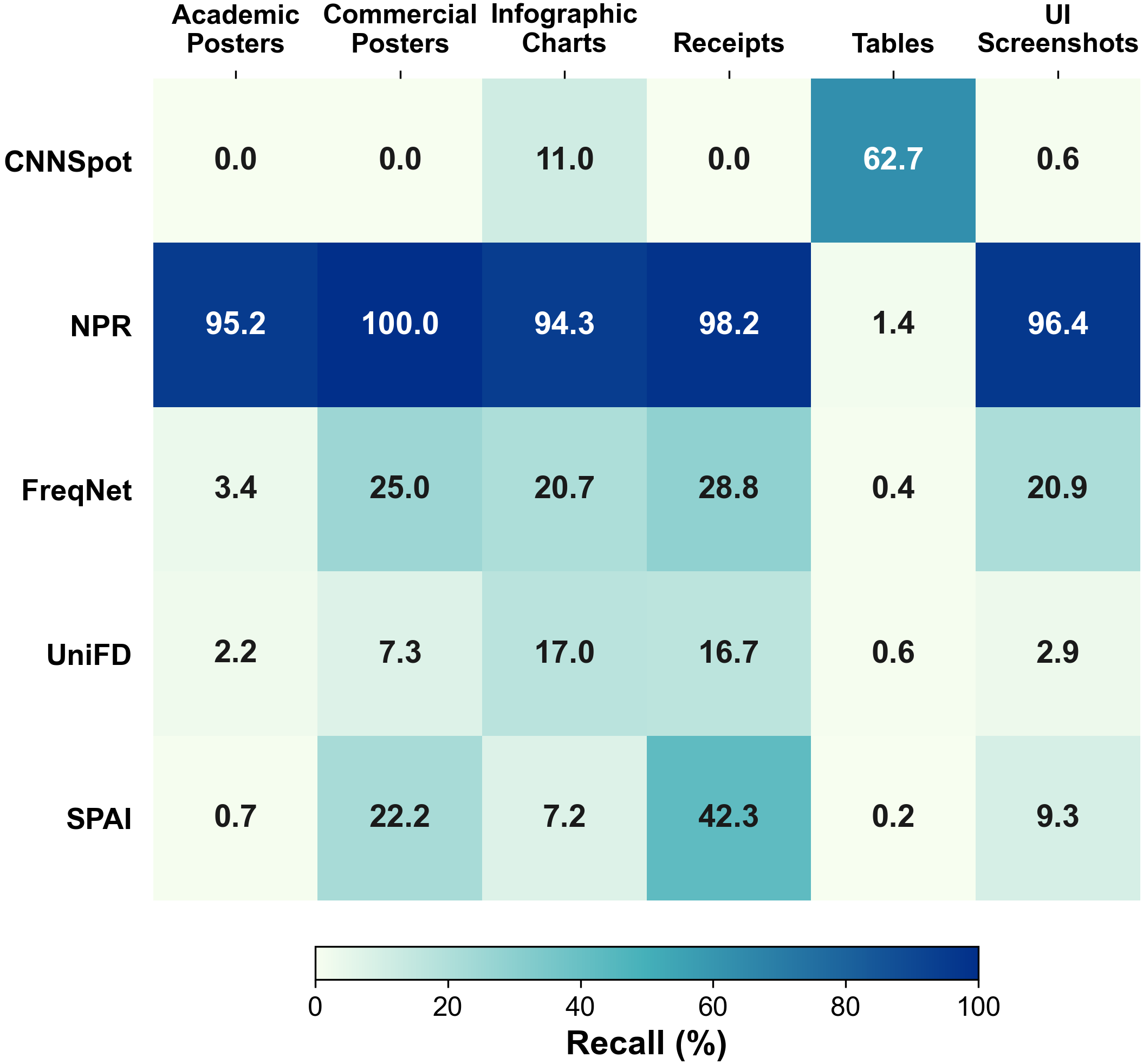}
    \vspace{-15pt}
    \captionof{figure}{Category-wise detection recall of 5 AIGI detectors on TextRich.}
    \label{fig:heatmap_recall}
\end{minipage}
The detectors were directly evaluated on these degraded images without additional fine-tuning. During inference, each image was evaluated exactly once and produced a single prediction. Classification decisions were generated using a threshold of 0.5 on the detector output score. No repeated sampling, voting strategy, or ensemble method was employed.

\begin{table}[!t]
\centering
\begin{tabularx}{\textwidth}{l*{6}{>{\centering\arraybackslash}X}}
\toprule
\textbf{Detector} 
& \textbf{TP} 
& \textbf{TN} 
& \textbf{Accuracy} 
& \textbf{Precision} 
& \textbf{Recall} 
& \textbf{F1} \\
\midrule
CNNSpot~\citep{wang2020cnngenerated} & 743  & 5926   & 0.5514 & 0.9253 & 0.1216 & 0.2150 \\
FreqNet~\citep{tan2024frequency}  & 1022 & 5936   & 0.5753 & 0.9534 & 0.1673 & 0.2846 \\
SPAI~\citep{karageorgiou2025any}       & 861  & 5746  & 0.5463 & 0.7820 & 0.1409 & 0.2388 \\
UniFD~\citep{ojha2024universalfakeimagedetectors}       & 484  & 5604  & 0.5033 & 0.5589 & 0.0792 & 0.1388 \\
NPR~\citep{tan2024rethinking}          & 4962 & 5797  & 0.8895 & 0.9633 & 0.8122 & 0.8814 \\
\bottomrule
\end{tabularx}

\caption{Performance of 5 AIGI detectors on TextRich.}
\label{tab:detector_results}
\end{table}

\subsection{Evaluation Metrics}
We evaluate detector performance using accuracy, precision, recall, and F1 score, treating AI-generated images as the positive class. In this setting, true positives (TP) denote generated images correctly classified as fake, while true negatives (TN) denote real images correctly classified as real. Precision measures how reliable the fake predictions are, while recall reflects how many generated images are successfully detected. 

\subsection{Detection Performance}
    
    \paragraph{Performance Overview.}
    ~\cref{tab:detector_results} summarizes the overall detection performance of all evaluated detectors. Among the six methods, NPR consistently outperforms all other detectors across all evaluation metrics, including accuracy, precision, recall, and F1-score. In contrast, UniFD consistently obtains the lowest scores, indicating limited effectiveness on the proposed dataset.
    Although CNNSpot, FreqNet, and SPAI obtain relatively high precision, their recall and F1-scores remain considerably lower than those of NPR. A closer examination of the TP and TN counts reveals that these detectors correctly classify most real images, with TN values close to the total number of real samples (5986), while identifying only a small fraction of AI-generated images. Even FreqNet, which records the highest TP count among the non-NPR methods, correctly detects only 1022 out of 6109 generated images.

These findings indicate that existing detectors exhibit a strong bias toward the real class on text-rich samples, correctly recognizing most real images while failing to detect a large proportion of AI-generated images. In contrast, the strong results achieved by NPR suggest that local neighboring pixel relationships introduced by generative up-sampling could be a promising technique for detecting text-rich images generated by modern large models.


\paragraph{Category-Wise Performance.}
To further understand which types of AI-generated images are most difficult to detect, ~\cref{fig:heatmap_recall} presents the category-wise recall for each method. Performance varies substantially across categories, with Academic Posters and Tables emerging as the most challenging. For Academic Posters, all detectors except NPR achieve recall values close to zero. The Table category remains particularly challenging, with NPR achieving a recall close to zero despite its strong overall performance, while the other detectors show similarly poor performance except CNNSpot, which stands out with a recall of over 62\% on the Table category despite showing relatively poor performance on most other categories. 
These results reveal that detector performance is highly dependent on the interaction between image structure and the underlying detection cues. While NPR benefits from low-level pixel artifacts introduced by generative up-sampling and achieves strong overall performance, highly structured categories such as tables remain challenging. The distinct behavior of CNNSpot further suggests that detector-specific artifacts do not consistently transfer across text-rich domains, highlighting the need for more robust detection strategies that account for diverse textual and structural patterns.

\begin{figure}[!t]
\centering
\begin{minipage}[t]{0.38\textwidth}
    \centering
    \includegraphics[width=\textwidth]{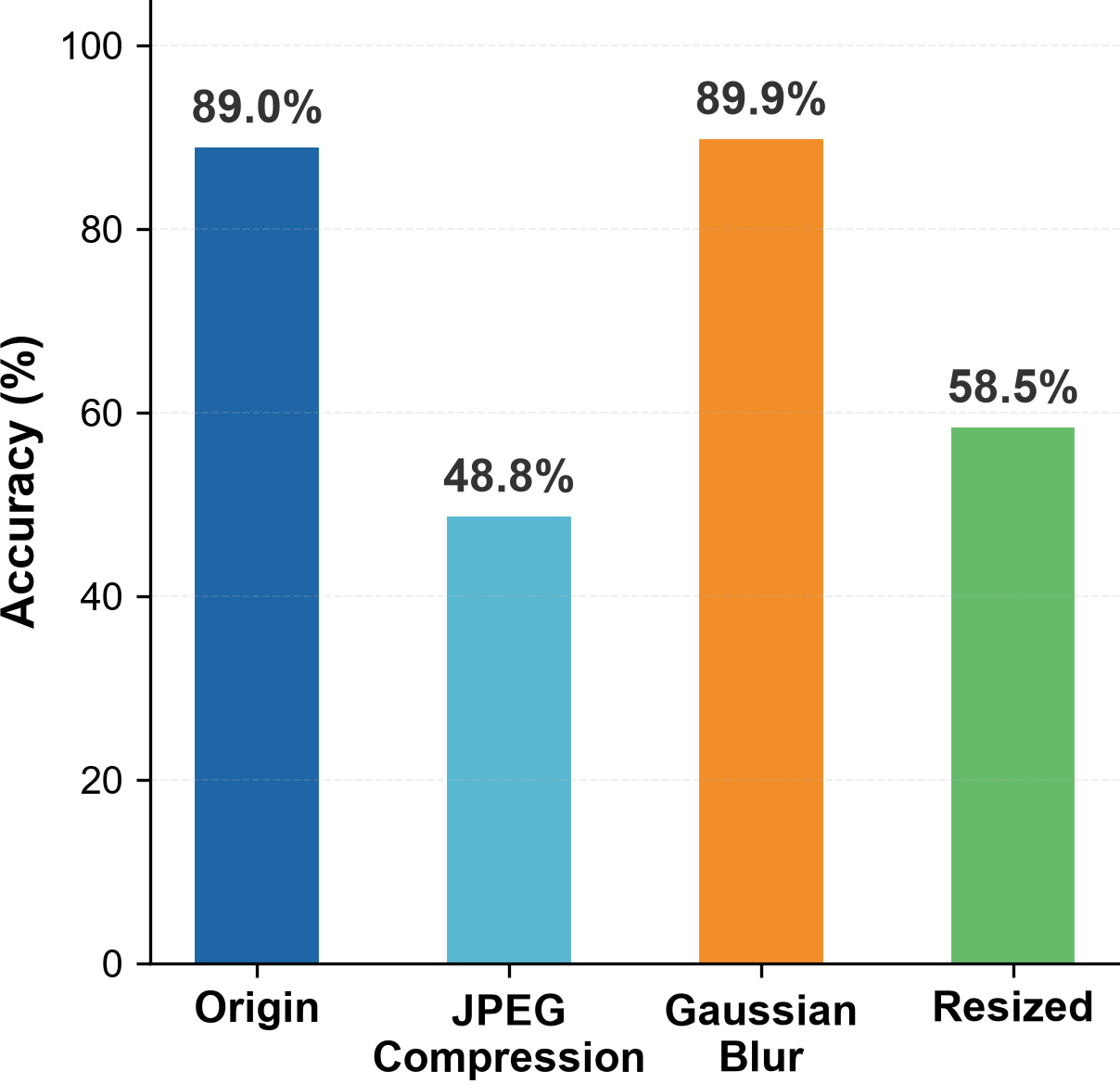}
    \vspace{-15pt}
    \caption{Overall accuracy of NPR under different degradation settings.}
    \label{fig:robustness_acc_bar}
\end{minipage}
\hfill
\begin{minipage}[t]{0.58\textwidth}
    \centering
    \includegraphics[width=\textwidth]{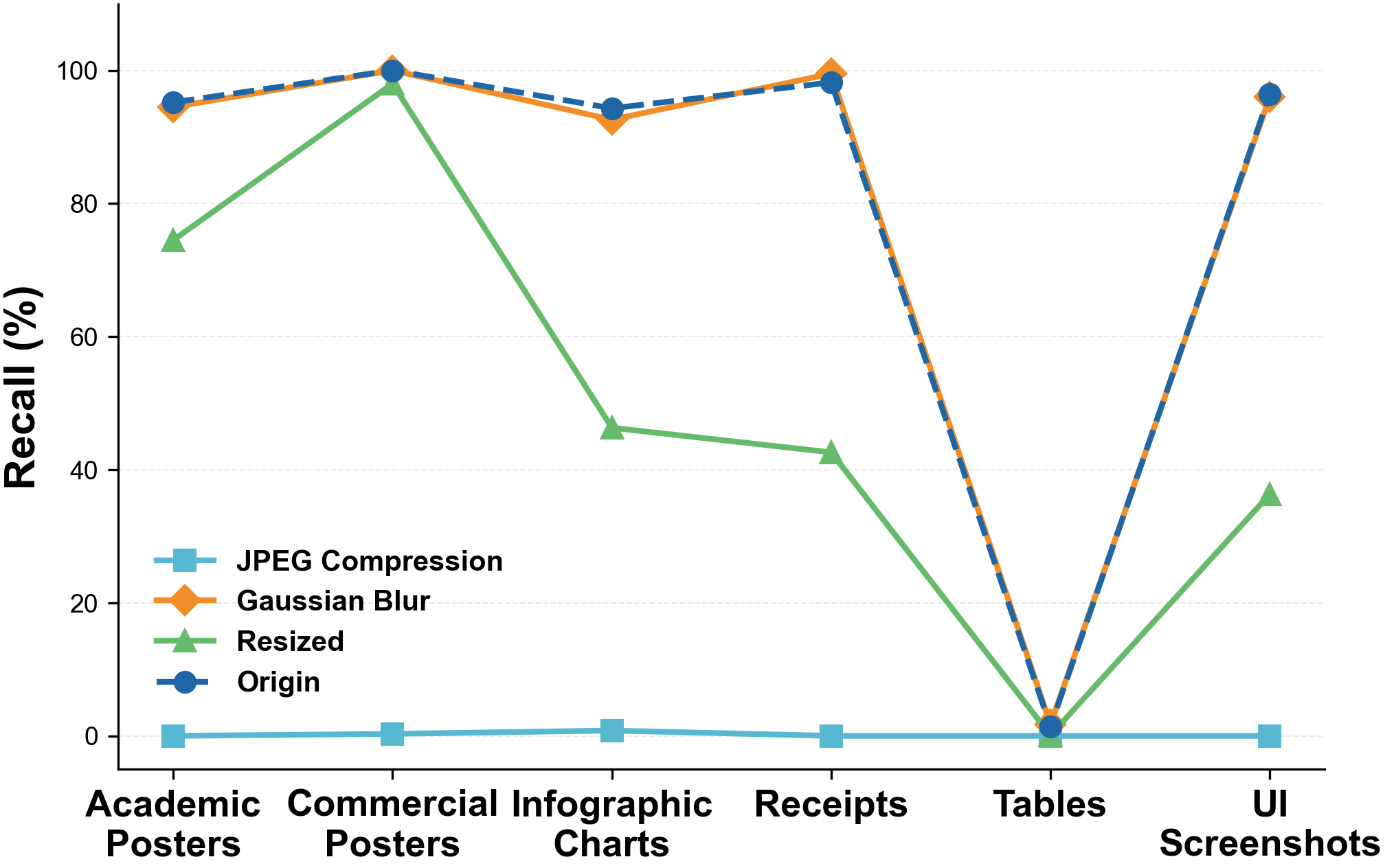}
    \vspace{-15pt}
    \caption{Category-wise detection recall of NPR under different image degradation settings.}
    \label{fig:robustness_recall}
\end{minipage}

\end{figure}

\subsection{Robustness Analysis}
Detector performance in practical settings may be affected by image post-processing operations such as resizing and compression. In our dataset, generated images are stored in PNG format, while some real images (e.g., UI screenshots and commercial posters) are collected in JPG or JPEG format. In addition, prior studies show that online platforms often apply further compression and resizing during image distribution~\citep{dang2023practical,hiney2015using}. Motivated by these observations, we conduct robustness experiments to evaluate the impact of such post-processing on detection performance. Since NPR achieves the best overall performance among the evaluated detectors, we select NPR as a representative strong baseline for this analysis and investigate whether its detection capability remains reliable under common image degradations.

As shown in ~\cref{fig:robustness_acc_bar}, Gaussian blur has almost no negative impact on detection performance, with the overall accuracy remaining nearly unchanged and even showing a slight improvement. In contrast, compression and resolution reduction, indicating that NPR relies, at least in part, on fine-grained image details for accurate prediction. A more detailed analysis is presented in ~\cref{fig:robustness_recall}, which reports the recall for each image category under different degradations. Among the three post-processing operations, JPEG compression has the most severe effect. The recall drops to nearly zero across all six categories, demonstrating that even moderate lossy compression can largely eliminate the features exploited by NPR. In comparison, Gaussian blur causes only minor variations across categories, while resolution reduction results in a moderate but relatively consistent decline in recall.

These results suggest that although NPR achieves excellent performance on the original dataset, its robustness to JPEG compression remains limited. This finding highlights the importance of evaluating AIGI detectors under realistic post-processing conditions and suggests that improving robustness to lossy compression remains an important direction for future research.

\begin{table}[!t]
\centering
\begin{tabularx}{0.98\textwidth}{l*{6}{>{\centering\arraybackslash}X}}
\toprule
\textbf{Category} 
& \textbf{TP} 
& \textbf{TN} 
& \textbf{Accuracy} 
& \textbf{Precision} 
& \textbf{Recall} 
& \textbf{F1} \\
\midrule
Academic Posters  & 597  & 945   & 0.7695 & 0.9156 & 0.5946 & 0.7210 \\
Commercial Posters & 773  & 673   & 0.7194 & 0.7027 & 0.7653 & 0.7327 \\
Infographic Charts & 634  & 913   & 0.7751 & 0.8842 & 0.6340 & 0.7385 \\
Receipts          & 927  & 950   & 0.9020 & 0.9626 & 0.8466 & 0.9009 \\
Tables            & 69   & 889   & 0.4790 & 0.3833 & 0.0690 & 0.1169 \\
UI Screenshots    & 614  & 995   & 0.8049 & 0.9935 & 0.6140 & 0.7590 \\
\midrule
\textbf{Overall} & 3614 & 5365  & 0.7427 & 0.8544 & 0.5916 & 0.6991 \\
\bottomrule
\end{tabularx}

\caption{Performance of Qwen3.5-27B across different categories on TextRich.}
\label{tab:qwen35_results}
\end{table}

\subsection{Multimodal Detectors}

We further conduct an exploratory evaluation using Qwen3.5-27B, an open-source vision-language model capable of understanding both visual and textual information~\citep{qwen3.5}. Unlike conventional detectors designed specifically for AIGI detection, Qwen3.5-27B represents a multimodal approach for general purposes. The model is prompted to classify each image as either real or AI-generated, while the category-wise and overall performance are summarized in \cref{tab:qwen35_results}. Overall, while the VLM falls short of the top-performing detector (NPR) from the previous evaluation. At the category level, the VLM exhibits relatively consistent performance across five of the six categories, while Receipt achieves the highest detection accuracy. In contrast, the Table category remains particularly challenging. Its accuracy drops to 47.90\%, with a recall of only 6.90\%, indicating that the model fails to identify most AI-generated table images and instead classifies them as real. These results indicate that vision-language models offer promising detection capabilities, but still struggle with text-rich images where precise structural alignment and layout understanding are required.

\section{Conclusion}
In this work, we introduce a multi-domain benchmark for detecting AI-generated text-rich images, consisting of 12,095 real and synthetic images across six representative categories generated with GPT-Image-2. Through comprehensive zero-shot evaluations, we show that existing AI-generated image detectors face significant challenges when applied to text-rich visual domains. Different detection paradigms exhibit distinct strengths and limitations, where local pixel relationship modeling can achieve strong overall performance but remain vulnerable to structured layouts and common post-processing operations, while multimodal vision-language models provide a promising complementary perspective yet struggle with highly organized text and layout structures. Through this benchmark, we provide a systematic evaluation platform for studying the challenges of AI-generated text-rich image detection and advancing future AI-generated image detection methods. We hope this work contributes to the development of more reliable and generalizable approaches for ensuring content authenticity in increasingly sophisticated multimodal generation environments.


%
%
%
%

\bibliographystyle{plainnat}
\bibliography{references}

@inproceedings{wang2020cnngenerated,
  title     = {CNN-generated images are surprisingly easy to spot... for now},
  author    = {Wang, Sheng-Yu and Wang, Oliver and Zhang, Richard and Owens, Andrew and Efros, Alexei A.},
  booktitle = {Proceedings of the IEEE/CVF Conference on Computer Vision and Pattern Recognition},
  year      = {2020}
}

@inproceedings{tan2024rethinking,
  title={Rethinking the up-sampling operations in cnn-based generative network for generalizable deepfake detection},
  author={Tan, Chuangchuang and Zhao, Yao and Wei, Shikui and Gu, Guanghua and Liu, Ping and Wei, Yunchao},
  booktitle={Proceedings of the IEEE/CVF conference on computer vision and pattern recognition},
  pages={28130--28139},
  year={2024}
}

@misc{ojha2024universalfakeimagedetectors,
      title={Towards Universal Fake Image Detectors that Generalize Across Generative Models}, 
      author={Utkarsh Ojha and Yuheng Li and Yong Jae Lee},
      year={2024},
      eprint={2302.10174},
      archivePrefix={arXiv},
      primaryClass={cs.CV},
      url={https://arxiv.org/abs/2302.10174}, 
}

@article{tan2024frequency,
  title={Frequency-aware deepfake detection: Improving generalizability through frequency space domain learning},
  author={Tan, Chuangchuang and Zhao, Yao and Wei, Shikui and Gu, Guanghua and Liu, Ping and Wei, Yunchao},
  journal={Proceedings of the AAAI Conference on Artificial Intelligence},
  volume={38},
  number={5},
  pages={5052--5060},
  year={2024}
}

@inproceedings{karageorgiou2025any,
  title={Any-resolution ai-generated image detection by spectral learning},
  author={Karageorgiou, Dimitrios and Papadopoulos, Symeon and Kompatsiaris, Ioannis and Gavves, Efstratios},
  booktitle={Proceedings of the Computer Vision and Pattern Recognition Conference},
  pages={18706--18717},
  year={2025}
}

@inproceedings{Yamaguchi_2021_ICCV,
    author = {Yamaguchi, Kota},
    title = {CanvasVAE: Learning To Generate Vector Graphic Documents},
    booktitle = {Proceedings of the IEEE/CVF International Conference on Computer Vision (ICCV)},
    month = {October},
    year = {2021},
    pages = {5481-5489}
}

@inproceedings{saxena2025postersum,
  title={Postersum: A multimodal benchmark for scientific poster summarization},
  author={Saxena, Rohit and Minervini, Pasquale and Keller, Frank},
  booktitle={Proceedings of the 14th International Joint Conference on Natural Language Processing and the 4th Conference of the Asia-Pacific Chapter of the Association for Computational Linguistics},
  pages={1828--1844},
  year={2025}
}

@inproceedings{huang2019icdar2019,
  title={Icdar2019 competition on scanned receipt ocr and information extraction},
  author={Huang, Zheng and Chen, Kai and He, Jianhua and Bai, Xiang and Karatzas, Dimosthenis and Lu, Shijian and Jawahar, CV},
  booktitle={2019 International Conference on Document Analysis and Recognition (ICDAR)},
  pages={1516--1520},
  year={2019},
  organization={IEEE}
}

@inproceedings{smock2022pubtables,
  title={PubTables-1M: Towards comprehensive table extraction from unstructured documents},
  author={Smock, Brandon and Pesala, Rohith and Abraham, Robin},
  booktitle={Proceedings of the IEEE/CVF Conference on Computer Vision and Pattern Recognition},
  pages={4634--4642},
  year={2022}
}

@inproceedings{leiva2020enrico,
  title={Enrico: A dataset for topic modeling of mobile UI designs},
  author={Leiva, Luis A and Hota, Asutosh and Oulasvirta, Antti},
  booktitle={22nd International Conference on Human-Computer Interaction with Mobile Devices and Services},
  pages={1--4},
  year={2020}
}

@article{li2025chartgalaxy,
  title={Chartgalaxy: A dataset for infographic chart understanding and generation},
  author={Li, Zhen and Li, Duan and Guo, Yukai and Guo, Xinyuan and Li, Bowen and Xiao, Lanxi and Qiao, Shenyu and Chen, Jiashu and Wu, Zijian and Zhang, Hui and others},
  journal={arXiv preprint arXiv:2505.18668},
  year={2025}
}

@inproceedings{dang2023practical,
  title={Practical analyses of how common social media platforms and photo storage services handle uploaded images},
  author={Dang-Nguyen, Duc-Tien and Sj{\o}en, Vegard Velle and Le, Dinh-Hai and Dao, Thien-Phu and Tran, Anh-Duy and Tran, Minh-Triet},
  booktitle={International conference on multimedia modeling},
  pages={164--176},
  year={2023},
  organization={Springer}
}

@inproceedings{hiney2015using,
  title={Using facebook for image steganography},
  author={Hiney, Jason and Dakve, Tejas and Szczypiorski, Krzysztof and Gaj, Kris},
  booktitle={2015 10th international conference on availability, reliability and security},
  pages={442--447},
  year={2015},
  organization={IEEE}
}

@inproceedings{NEURIPS2023_1df4afb0,
 author = {Chen, Jingye and Huang, Yupan and Lv, Tengchao and Cui, Lei and Chen, Qifeng and Wei, Furu},
 booktitle = {Advances in Neural Information Processing Systems},
 editor = {A. Oh and T. Naumann and A. Globerson and K. Saenko and M. Hardt and S. Levine},
 pages = {9353--9387},
 publisher = {Curran Associates, Inc.},
 title = {TextDiffuser: Diffusion Models as Text Painters},
 volume = {36},
 year = {2023}
}

@inproceedings{ICLR2024_081b0806,
 author = {Podell, Dustin and English, Zion and Lacey, Kyle and Blattmann, Andreas and Dockhorn, Tim and M\"{u}ller, Jonas and Penna, Joe and Rombach, Robin},
 booktitle = {International Conference on Learning Representations},
 editor = {B. Kim and Y. Yue and S. Chaudhuri and K. Fragkiadaki and M. Khan and Y. Sun},
 pages = {1862--1874},
 title = {SDXL: Improving Latent Diffusion Models for High-Resolution Image Synthesis},
 volume = {2024},
 year = {2024}
}

@misc{OpenAI2026Images2,
  author = {{OpenAI}},
  title = {Introducing {ChatGPT} {Images} 2.0},
  year = {2026},
  date = {2026-04-21},
  url = {https://openai.com/index/introducing-chatgpt-images-2-0/},
  urldate = {2026-05-25},
  organization = {OpenAI},
  note = {Accessed: 2026-05-25}
}

@misc{openai2026gpt54,
  author = {OpenAI},
  title = {Introducing GPT‑5.4},
  year = {2026},
  date = {2026-03-05},
  url = {https://openai.com/index/introducing-gpt-5-4/},
  urldate = {2026-07-18},
  organization = {OpenAI},
  note = {Accessed: July 2026}
}

@misc{qwen3.5,
    title  = {{Qwen3.5}: Towards Native Multimodal Agents},
    author = {{Qwen Team}},
    month  = {February},
    year   = {2026},
    url    = {https://qwen.ai/blog?id=qwen3.5}
}

@misc{hong2024wildfakelargescalechallengingdataset,
      title={WildFake: A Large-scale Challenging Dataset for AI-Generated Images Detection}, 
      author={Yan Hong and Jianfu Zhang},
      year={2024},
      eprint={2402.11843},
      archivePrefix={arXiv},
      primaryClass={cs.CV},
      url={https://arxiv.org/abs/2402.11843}, 
}

@inproceedings{NEURIPS2023_f4d4a021,
 author = {Zhu, Mingjian and Chen, Hanting and YAN, Qiangyu and Huang, Xudong and Lin, Guanyu and Li, Wei and Tu, Zhijun and Hu, Hailin and Hu, Jie and Wang, Yunhe},
 booktitle = {Advances in Neural Information Processing Systems},
 editor = {A. Oh and T. Naumann and A. Globerson and K. Saenko and M. Hardt and S. Levine},
 pages = {77771--77782},
 publisher = {Curran Associates, Inc.},
 title = {GenImage: A Million-Scale Benchmark for Detecting AI-Generated Image},
 volume = {36},
 year = {2023}
}

@inproceedings{10222083,
  author={Rahman, Md Awsafur and Paul, Bishmoy and Sarker, Najibul Haque and Hakim, Zaber Ibn Abdul and Fattah, Shaikh Anowarul},
  booktitle={2023 IEEE International Conference on Image Processing (ICIP)}, 
  title={Artifact: A Large-Scale Dataset With Artificial And Factual Images For Generalizable And Robust Synthetic Image Detection}, 
  year={2023},
  volume={},
  number={},
  pages={2200-2204},
  doi={10.1109/ICIP49359.2023.10222083}
}

@inproceedings{10.1007/978-3-031-19815-1_13,
author="Wang, Yuxin
and Xie, Hongtao
and Xing, Mengting
and Wang, Jing
and Zhu, Shenggao
and Zhang, Yongdong",
editor="Avidan, Shai
and Brostow, Gabriel
and Ciss{\'e}, Moustapha
and Farinella, Giovanni Maria
and Hassner, Tal",
title="Detecting Tampered Scene Text in the Wild",
booktitle="Computer Vision -- ECCV 2022",
year="2022",
publisher="Springer Nature Switzerland",
address="Cham",
pages="215--232",
isbn="978-3-031-19815-1"
}

@article{qu2025revisiting,
  title={Revisiting tampered scene text detection in the era of generative AI},
  author={Qu, Chenfan and Zhong, Yiwu and Guo, Fengjun and Jin, Lianwen},
  journal={Proceedings of the AAAI Conference on Artificial Intelligence},
  volume={39},
  number={1},
  pages={694--702},
  year={2025}
}

@misc{wu2026aiforgedocbenchmarkdetectingaiforged,
  title={AIForge-Doc: A Benchmark for Detecting AI-Forged Tampering in Financial and Form Documents}, 
  author={Jiaqi Wu and Yuchen Zhou and Muduo Xu and Zisheng Liang and Simiao Ren and Jiayu Xue and Meige Yang and Siying Chen and Jingheng Huan},
  year={2026},
  eprint={2602.20569},
  archivePrefix={arXiv},
  primaryClass={cs.CV},
  url={https://arxiv.org/abs/2602.20569}, 
}

@misc{zhang2026gpt4oreceiptdatasethumanstudy,
      title={GPT4o-Receipt: A Dataset and Human Study for AI-Generated Document Forensics}, 
      author={Yan Zhang and Simiao Ren and Ankit Raj and En Wei and Dennis Ng and Alex Shen and Jiayu Xue and Yuxin Zhang and Evelyn Marotta},
      year={2026},
      eprint={2603.11442},
      archivePrefix={arXiv},
      primaryClass={cs.AI},
      url={https://arxiv.org/abs/2603.11442}, 
}

@misc{hu2026scifigdetectbenchmarkaigeneratedscientific,
      title={SciFigDetect: A Benchmark for AI-Generated Scientific Figure Detection}, 
      author={You Hu and Chenzhuo Zhao and Changfa Mo and Haotian Liu and Xiaobai Li},
      year={2026},
      eprint={2604.08211},
      archivePrefix={arXiv},
      primaryClass={cs.CV},
      url={https://arxiv.org/abs/2604.08211}, 
}

@article{fu2025multimodal,
  title={Multimodal large language models for text-rich image understanding: A comprehensive review},
  author={Fu, Pei and Guan, Tongkun and Wang, Zining and Guo, Zhentao and Duan, Chen and Sun, Hao and Chen, Boming and Jiang, Qianyi and Ma, Jiayao and Zhou, Kai and others},
  journal={Findings of the Association for Computational Linguistics: ACL 2025},
  pages={19941--19958},
  year={2025}
}

@article{ding2026deep,
  title={Deep learning based visually rich document content understanding: A survey},
  author={Ding, Yihao and Han, Soyeon Caren and Lee, Jean and Hovy, Eduard},
  journal={Artificial Intelligence Review},
  year={2026},
  publisher={Springer}
}

\end{document}